\pdfoutput=1

\documentclass[11pt]{article}

\usepackage{coling}

\usepackage{times}
\usepackage{latexsym}

\usepackage[T1]{fontenc}

\usepackage[utf8]{inputenc}

\usepackage{microtype}

\usepackage{inconsolata}

\usepackage{graphicx}

\usepackage{pgfplots}
\usepackage{pgf} 
\pgfplotsset{compat=1.16}

\usepackage{algorithm}
\usepackage{algorithmic}

\usepackage[most]{tcolorbox}
\usepackage{enumitem}
\usepackage{multirow}

\title{Enhancing Future Link Prediction in Quantum Computing Semantic Networks through LLM-Initiated Node Features}

\author{
  \textbf{Gilchan Park\textsuperscript{*}},
  \textbf{Paul Baity},
  \textbf{Byung-Jun Yoon},
  \textbf{Adolfy Hoisie}
\\
  Brookhaven National Laboratory, Computational Science Initiative
\\
  Upton, New York, USA
\\
  \small{
    \textbf{*Correspondence:} \href{mailto:gpark@bnl.gov}{gpark@bnl.gov}
  }
}

\begin{document}
\maketitle
\begin{abstract}
Quantum computing is rapidly evolving in both physics and computer science, offering the potential to solve complex problems and accelerate computational processes. The development of quantum chips necessitates understanding the correlations among diverse experimental conditions. Semantic networks built on scientific literature, representing meaningful relationships between concepts, have been used across various domains to identify knowledge gaps and novel concept combinations. Neural network-based approaches have shown promise in link prediction within these networks. This study proposes initializing node features using LLMs to enhance node representations for link prediction tasks in graph neural networks. LLMs can provide rich descriptions, reducing the need for manual feature creation and lowering costs. Our method, evaluated using various link prediction models on a quantum computing semantic network, demonstrated efficacy compared to traditional node embedding techniques. The code and data are available at: \tt{\url{https://github.com/boxorange/QC-LinkPrediction}}
\end{abstract}

\section{Introduction}

Quantum computing is an active area of research in both physics and computer science, due to its potential to solve complex quantum physics problems and significantly accelerate certain computational processes \citep{Shor1997,Montanaro2016,Arute2019}. However, the current limitations of hardware hinder the practical application of quantum computers \citep{Krantz2019,Kjaergaard2020}, and the further development of robust quantum processors involves an increasingly wide range of conditions \citep{Huang2021,Martinis2021}, material characteristics \cite{Murray2021,Place2021}, and physical phenomena. Understanding the correlations among these variables and predicting their potential interconnections in the future is crucial for experimental progress. Scientific literature serves as a vital resource for acquiring this knowledge, as it encompasses a vast array of research work. 

A semantic network represents meaningful relationships between concepts, and researchers constructed a semantic network based on co-occurring concepts from scientific literature, utilizing it to identify knowledge gaps, missing connections between concepts, and novel combinations not previously considered \citep{rzhetsky2015choosing, krenn2020predicting}. In recent years, Graph Neural Networks (GNNs) demonstrated promising predictive capabilities for link prediction within the graph forms of semantic networks \citep{zhang2018link, NEURIPS2023_0be50b45}. A significant challenge in creating semantic networks is the provision of sufficient initial features for nodes within a graph. In many real-world graph datasets, node features are often either missing or insufficient, potentially hindering link prediction models for effective learning and prediction \citep{zhao2017leveraging}.

This study aims to initialize node features using Large Language Models (LLMs). LLMs have demonstrated exceptional performance across various question-answering tasks and information retrieval systems in zero-shot conditions \citep{kamalloo2023evaluating, LLM4IRSurvey}, significantly improving text embeddings \citep{wang2023improving}. These embeddings serve as initial node representations for link prediction tasks in GNNs. The rationale behind this approach is that LLMs, trained on extensive datasets from diverse literature and online sources, can provide rich descriptions of relevant concepts. This method enhances the feature set available for GNN training and reduces the reliance on human-curated feature creation. Additionally, it has the potential to produce more reliable node representations compared to traditional connectivity-based embeddings, particularly when connectivity data is lacking. In cold-start link prediction problems \citep{sedhain2014social, zhang2015collective, tang2022cold}, where nodes lack edges, informative node features become critical \citep{zhao2017leveraging}, contributing to the generation of structural information and facilitating link formation \citep{mullerattending}. Our approach offers a straightforward yet impactful method for node feature initialization using LLMs, without the need for external resources or dependency on graph structure. We evaluated this method through various link prediction models, conducting a comparative analysis with widely-used node embedding techniques within a quantum computing semantic network.

\section{Creation of Semantic Network for Quantum Computing}

The construction of a concept network from scratch necessitates significant human resources and time. As an alternative, we utilize the pre-existing semantic network of quantum physics. \textsc{SemNet} \citep{krenn2020predicting} was developed to forecast future research trends in quantum physics. It is based on scientific publications in the field of quantum physics and includes a list of human-compiled quantum physics concepts derived from books and Wikipedia. In \textsc{SemNet}, nodes represent physical concepts, and edges indicate the co-occurrence of two concepts in an article's title or abstract.

Our domain experts in quantum physics have scrutinized the list of concepts, selecting those that have a high correlation with quantum computing research. Certain concepts were revised, and additional ones that appeared to be pertinent to quantum computing research, such as "quasiparticle poisoning," were incorporated. This process resulted in a compilation of 3,001 quantum computing concepts.

Given the lack of a corpus upon which \textsc{SemNet} was created, we scraped all articles under the quantum physics category on arXiv that contained at least a pair of quantum computing concepts. This resulted in a total of 136,122 papers spanning from 2007 to 2024 (as of June 15, 2024). Appendix~\ref{sec:appendix-pub-stat} illustrates the number of quantum computing papers published on arXiv over time. 

\begin{figure*}[t]
  \includegraphics[width=\linewidth, height=7.5cm]{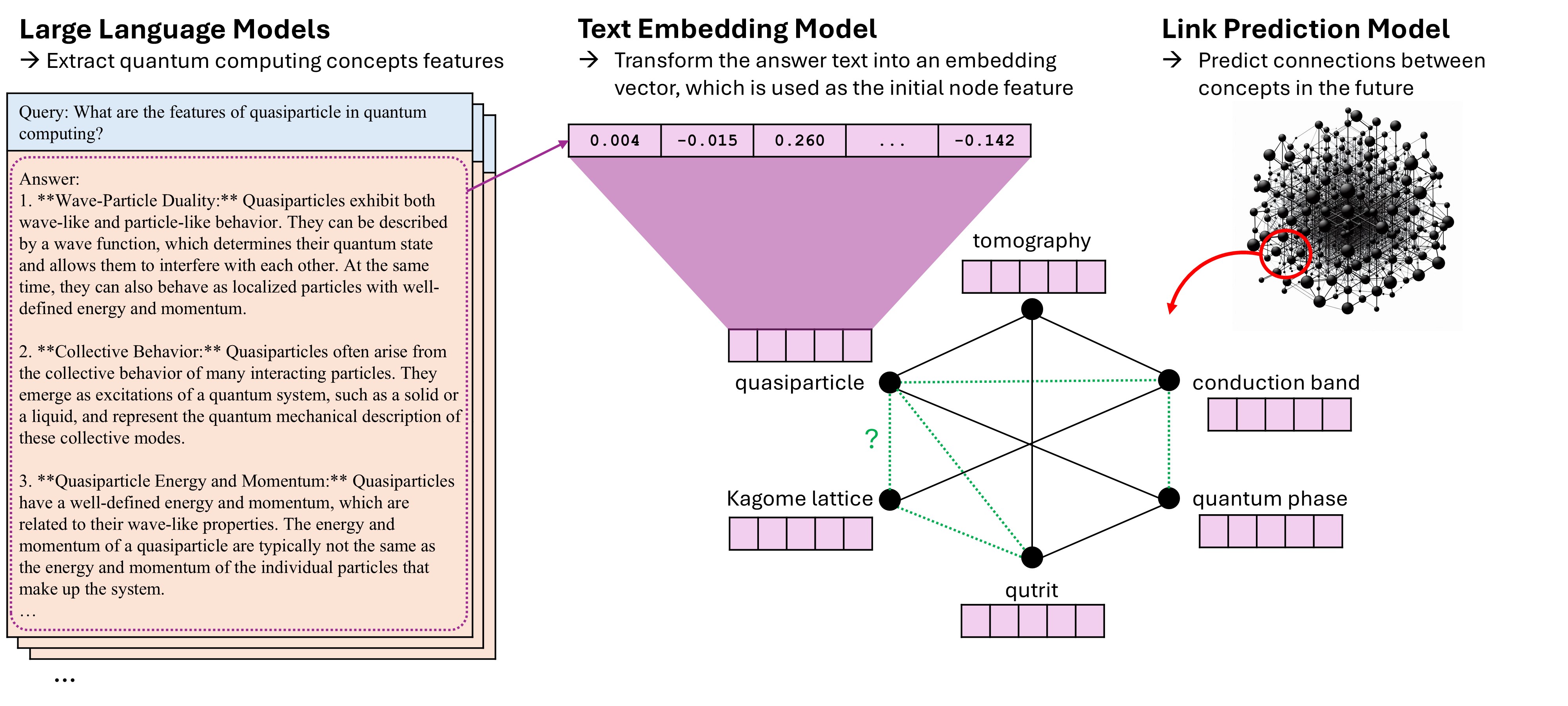}
  \caption {The overview of future link predictions in the quantum computing semantic network using LLM-generated initial node features. In the example graph, solid lines indicate past established connections, while dotted lines represent a subset of potential future connections to be predicted by the model for relevance.}
  \label{fig:model}
\end{figure*}

\section{LLM-powered Concept Feature Embedding}

We derived quantum computing concept features from LLMs using a question-answering prompt. For this purpose, we employed a selection of state-of-the-art LLMs, specifically Gemini-1.0-pro \citep{team2023gemini}, Mixtral \citep{jiang2024mixtral}, and LLaMA 3 \citep{metallama3}. The following example illustrates a query prompt and the corresponding answer generated by an LLM.

\begin{tcolorbox}[halign=left, fontupper=\footnotesize, fontlower=\footnotesize, boxsep=0.1pt]
\textbf{Query:}  What are the features of quasiparticle in quantum computing?\\~\\
\textbf{Answer:} 

1. **Wave-Particle Duality:** Quasiparticles exhibit both wave-like and particle-like behavior. They can be described by a wave function, which determines their quantum state and allows them to interfere with each other. At the same time, they can also behave as localized particles with well-defined energy and momentum.

2. **Collective Behavior:** Quasiparticles often arise from the collective behavior of many interacting particles. They emerge as excitations of a quantum system, such as a solid or a liquid, and represent the quantum mechanical description of these collective modes.

3. **Quasiparticle Energy and Momentum:** Quasiparticles have a well-defined energy and momentum, which are related to their wave-like properties. The energy and momentum of a quasiparticle are typically not the same as the energy and momentum of the individual particles that make up the system.

...
\end{tcolorbox}

We utilized the Google AI Text Embedding model \citep{google_text_embedding}, text-embedding-004, to generate text embeddings from the concept features produced by the LLMs. Figure~\ref{fig:model} illustrates the node features derived from the LLMs, which are used in the link prediction models.

\section{Experiments}

We evaluated the initial node representations generated by LLMs in a transductive experimental setup, where models predict edges between existing nodes in the graph. The models were trained on historical concept connections and tasked with predicting future, unknown connections. The evaluation was performed using various link prediction algorithms.

\subsection{Experiment Setup}

\vspace{2mm}
{\setlength{\parindent}{0cm}\textbf{Dataset} \hspace{0cm}
An undirected homogeneous (binary) graph, also known as a single relational graph, was constructed for the purpose of link prediction. The dataset was divided into three subsets based on specific time intervals, a common approach for time series data in link prediction tasks (chronological splitting). The training set encompassed the period from 2007 to 2021 and included 428,079 edges. The validation set corresponded to the year 2022 and contained 25,011 edges. The test set covered the most recent year, from 2023 to 2024, comprising 50,063 edges. The dataset was distributed with an approximate ratio of 85:5:10 for training, validation, and testing, respectively. 
}

\vspace{2mm}
{\setlength{\parindent}{0cm}\textbf{Link Prediction Models \& Baselines} \hspace{0.2cm}
We conducted an evaluation of the proposed node features across three classes of link prediction models: (1) Multi-layer Perceptron (MLP), (2) message-passing mechanism-based GNNs including GraphSAGE \citep{hamilton2017inductive}, GCN \citep{kipf2016semi}, and GAE \citep{kipf2016variational}, and (3) GNNs with pair-wise information methods, specifically NCN \citep{wangneural} and BUDDY \citep{chamberlain2023graph} that leverage common neighbor information and sub-graph features to further capture the relation between the nodes for potential links respectively. The selection of these GNN models was based on their high ranking in recent comprehensive evaluations of link prediction methods \citep{NEURIPS2023_0be50b45}.

We conducted a comparative analysis of the embeddings generated by LLMs with those produced by widely recognized node embedding methods, including DeepWalk \citep{perozzi2014deepwalk}, LINE \citep{tang2015line}, and node2vec \citep{grover2016node2vec}. These methods were employed in previous studies comparing node feature initialization techniques \citep{duong2019node, zhu2021neural, berahmand2021modified, cui2022positional}.
}

\begin{table*}
  \centering
  \renewcommand{\arraystretch}{1.3} 
  \small
  \begin{tabular}{l|cccccc}
    \hline
    \multirow{2}{*}{\textbf{Node Embedding}} & \multicolumn{2}{c}{\textbf{MLP}} & \multicolumn{2}{c}{\textbf{GCN}} & \multicolumn{2}{c}{\textbf{GraphSAGE}} \\
    & AUROC & AP & AUROC & AP & AUROC & AP \\
    \hline
    DeepWalk & 82.48 ± 0.19 & 80.72 ± 0.12 & 88.98 ± 0.14 & 87.80 ± 0.17 & 86.40 ± 0.21 & 84.51 ± 0.24 \\
    LINE & 83.92 ± 0.48 & 81.94 ± 0.52 & 86.95 ± 0.03 & 85.51 ± 0.02 & 83.25 ± 2.62 & 80.48 ± 4.11 \\
    node2vec & 84.82 ± 0.20 & 82.75 ± 0.25 & 88.27 ± 0.18 & 86.76 ± 0.28 & 87.06 ± 0.14 & 85.31 ± 0.20 \\
    \hline
    Gemini-1.0-pro & 86.56 ± 0.20 & 84.62 ± 0.26 & \textbf{89.63 ± 0.05} & \textbf{88.39 ± 0.05} & 88.79 ± 0.12 & 87.33 ± 0.14 \\
    LLaMA3 (70B) & 86.15 ± 0.24 & 84.18 ± 0.29 & 89.52 ± 0.06 & 88.29 ± 0.07 & 88.67 ± 0.09 & 87.16 ± 0.11 \\
    Mixtral-8x7B (46B) & \textbf{87.02 ± 0.22} & \textbf{85.14 ± 0.27} & 89.61 ± 0.08 & 88.38 ± 0.11 & \textbf{88.87 ± 0.10} & \textbf{87.45 ± 0.18} \\
    \hline
  \end{tabular}
  \\[0.3cm]
  \begin{tabular}{l|cccccc}
    \hline
    \multirow{2}{*}{\textbf{Node Embedding}} & \multicolumn{2}{c}{\textbf{GAE}} & \multicolumn{2}{c}{\textbf{NCN}} & \multicolumn{2}{c}{\textbf{BUDDY}} \\
    & AUROC & AP & AUROC & AP & AUROC & AP \\
    \hline
    DeepWalk & 86.18 ± 0.10 & 84.39 ± 0.10 & 88.92 ± 0.17 & 87.52 ± 0.21 & 87.84 ± 0.07 & 86.31 ± 0.10 \\
    LINE & 86.56 ± 0.01 & 85.13 ± 0.01 & 88.83 ± 0.06 & 87.42 ± 0.08 & 87.63 ± 0.02 & 86.13 ± 0.02 \\
    node2vec & 81.89 ± 0.44 & 80.12 ± 0.43 & 88.98 ± 0.12 & \textbf{87.66 ± 0.15} & 88.55 ± 0.08 & 87.18 ± 0.08 \\
    \hline
    Gemini-1.0-pro & \textbf{87.27 ± 0.12} & \textbf{85.35 ± 0.13} & \textbf{89.07 ± 0.25} & 87.46 ± 0.28 & 88.79 ± 0.07 & 87.35 ± 0.09 \\
    LLaMA3 (70B) & 86.89 ± 0.11 & 85.03 ± 0.11 & 88.90 ± 0.24 & 87.33 ± 0.27 & 88.55 ± 0.04 & 87.09 ± 0.06 \\
    Mixtral-8x7B (46B) & 86.77 ± 0.15 & 84.91 ± 0.14 & 88.99 ± 0.27 & 87.44 ± 0.25 & \textbf{88.87 ± 0.06} & \textbf{87.40 ± 0.08} \\
    \hline
  \end{tabular}
  \caption{Comparison of LLM-generated node embeddings with other node embeddings in link prediction methods on a homogeneous, undirected graph representing quantum computing concept relations in a transductive setting.}
  \label{table:main}
\end{table*}

\vspace{2mm}
{\setlength{\parindent}{0cm}\textbf{Implementation Details \& Evaluation Setting} \hspace{0.2cm}
The experiments were conducted using 4$\times$NVIDIA A100 80GB GPUs. Specifically, the Mixtral-8x7B-Instruct (46B) and LLaMA-3 (70B) models were executed on 4$\times$NVIDIA A100 80GB GPUs to generate concept features. All link prediction methods were performed on a single NVIDIA A100 80GB GPU. The Google Gemini-1.0-pro and text-embedding-004 models were accessed via the Gemini APIs. The maximum number of generated tokens per query was set to 512 for all models, and the default embedding size of 768, as produced by the Google text embedding model, was employed. To maintain consistency, node embeddings of size 768 were also generated using the baseline methods. In terms of latency, the Gemini-pro model required approximately 3 hours, the Mixtral model 6 hours, and the LLaMA-3 (70B) model 8 hours to generate features related to quantum computing concepts. The text embedding process was completed in less than 2 minutes.
\\
We followed the hyper-parameter ranges for the models employed in the comprehensive link prediction evaluation \citep{NEURIPS2023_0be50b45}. For the model evaluation, we measured the area under the receiver operating characteristic curve (AUROC) and average precision (AP), which are commonly used metrics for link prediction tasks in homogeneous graphs. These metrics provide a robust and comprehensive assessment of the model's performance \citep{yang2015evaluating,zhu2021neural}. Each experiment was repeated ten times with different random seeds to ensure the reliability of the results.
}

\subsection{Node Embedding Comparison Results}
Table~\ref{table:main} presents the link prediction results using baseline and LLM-powered node embeddings. The majority of models initialized with LLM-generated embeddings demonstrated higher performance than their baseline counterparts. Notably, the LLM-generated features resulted in more significant improvements in MLP and message passing GNNs (GCN, GraphSAGE, GAE) than in the GNN with pair-wise information methods (NCN, BUDDY). This can be attributed to the fact that MLP and GNNs relying on message passing mechanisms are generally more impacted by the initial node embeddings compared to those models that incorporate additional link specific information. Message passing aggregates information from a node's neighborhood, and if the initial embedding already captures substantial information, it can have a stronger influence on the final embedding.

We further compared node feature initialization methods on isolated (zero-degree) nodes, which pose significant challenges for GNNs \citep{ahn2021variational, zanardini2024introducing}. We identified 30 isolated nodes in the training data and 1,382 connections to these in the test data. The evaluation results are presented in Appendix~\ref{sec:appendix-eval-iso}. Although the baseline methods exhibited higher performance in certain instances, particularly with the GCN in conjunction with LINE, the representations produced by LLMs were generally more effective in identifying previously unseen connections to isolated nodes. Furthermore, they yielded a more consistent performance in comparison to the baseline methodologies. For these isolated nodes, the content of the node features is crucial for the link prediction task due to the absence of connectivity information. The baseline models typically generate node embeddings based on the connectivity information of a graph, which may result in inadequate embeddings for isolated nodes. In contrast, embeddings generated by LLMs are robust against the absence of link connectivity information and can thus produce reliable representations for isolated nodes. 

\subsection{Merging LLM embeddings}
We conducted an evaluation incorporating features from various models. To merge conceptual features derived from multiple LLMs, we employed mean- and max-pooling on the embeddings, and we extracted concise conceptual features from the outputs of three LLMs using the Gemini-pro model. The prompt used was: "Summarize this text about the features of \{\textsc{keyword}\}. Text: \{\textsc{concatenated features from the LLMs}\}". Furthermore, we evaluated a method to select the optimal response for each query from three distinct models, utilizing LLM-Blender \citep{jiang2023llm}. This ensembling framework chooses the top score answer from multiple LLMs through a specialized pairwise comparison technique and a generative fusion module. These merged embeddings were compared against the leading LLM node embeddings across different models, including MLP, GCN (top performance among message-passing models), and NCN (the highest scorer in GNNs with pairwise information). Table~\ref{table:merging} presents the comparative analysis. The findings suggest that the use of simple mean or max pooling techniques can effectively merge embeddings, rendering more complex methods unnecessary.

\begin{table}
  \centering
  \renewcommand{\arraystretch}{1.3} 
  \small
  \begin{tabular}{ll|cc}
    \hline
    \textbf{Models} & \textbf{Embedding} &\textbf{AUROC} & \textbf{AP}\\
    \hline
    \multirow{5}{*}{MLP} 
    & Mixtral & 87.02 ± 0.22 & 85.14 ± 0.27 \\
    & Mean pool & 86.95 ± 0.17 & 84.96 ± 0.19 \\
    & Max pool & \textbf{87.06 ± 0.18} & \textbf{85.16 ± 0.19} \\
    & LLM-Blender & 86.71 ± 0.17 & 84.70 ± 0.17 \\
    & Summarized & 86.72 ± 0.14 & 84.81 ± 0.21 \\
    \hline
    \multirow{5}{*}{GCN}
    & Gemini Pro & 89.63 ± 0.05 & 88.39 ± 0.05 \\
    & Mean pool & \textbf{89.68 ± 0.08} & \textbf{88.45 ± 0.09} \\
    & Max pool & 89.58 ± 0.06 & 88.33 ± 0.09 \\
    & LLM-Blender & 89.60 ± 0.04 & 88.37 ± 0.08 \\
    & Summarized & 89.59 ± 0.06 & 88.37 ± 0.08 \\
    \hline
    \multirow{5}{*}{NCN}
    & Gemini Pro & 89.07 ± 0.25 & 87.46 ± 0.28 \\
    & Mean pool & \textbf{89.08 ± 0.22} & \textbf{87.64 ± 0.23} \\
    & Max pool & 88.96 ± 0.24 & 87.47 ± 0.25 \\
    & LLM-Blender & 88.96 ± 0.22 & 87.40 ± 0.20 \\
    & Summarized & 88.84 ± 0.23 & 87.29 ± 0.25 \\
    \hline
  \end{tabular}
  \caption{Comparison of different merging methods of LLM-generated feature embeddings.}
  \label{table:merging}
\end{table}

\subsection{Time Decayed Embedding}
Additionally, we incorporated time-decay information from the dataset. In future trend predictions involving time series data, time-decayed information can be important for maintaining the relevance of recent data and highlighting recent changes. To leverage the time-decay information of concept pairs over time, we developed time-decayed node representations based on co-occurrence matrices by year. These representations can be integrated with the LLM feature embeddings to enhance the model's capabilities. The time-decayed embeddings serve as optional auxiliary data in time series analyses, as they cannot function as standalone embeddings due to the potential lack of connections between concepts, which would result in non-informative embeddings. A comprehensive explanation of the time-decayed embedding generation process is available in the Appendix~\ref{sec:appendix-td}. Table~\ref{table:td} shows the performance of MLP, GCN, and NCN models with concatenated node embeddings with LLM features and time-decayed representations. The incorporation of time-decayed information in the node representation enhanced all models predictive capability, with the MLP model demonstrating the greatest improvement.

\begin{table}
  \centering
  \renewcommand{\arraystretch}{1.3} 
  \small
  \begin{tabular}{ll|cc}
    \hline
    \textbf{Models} & \textbf{Embedding} &\textbf{AUROC} & \textbf{AP}\\
    \hline
    \multirow{6}{*}{MLP} 
    & Gemini & 86.56 ± 0.20 & 84.62 ± 0.26 \\
    & Gemini + TD & 87.85 ± 0.24 & 86.25 ± 0.35 \\
    & LLaMA3 & 86.15 ± 0.24 & 84.18 ± 0.29 \\
    & LLaMA3 + TD & 87.26 ± 0.20 & 85.62 ± 0.27 \\
    & Mixtral & 87.02 ± 0.22 & 85.14 ± 0.27 \\
    & Mixtral + TD & \textbf{87.94 ± 0.28} & \textbf{86.33 ± 0.32} \\
    \hline
    \multirow{6}{*}{GCN}
    & Gemini & 89.63 ± 0.05 & 88.39 ± 0.05 \\
    & Gemini + TD & 89.65 ± 0.09 & 88.46 ± 0.09 \\
    & LLaMA3 & 89.52 ± 0.06 & 88.29 ± 0.07 \\
    & LLaMA3 + TD & 89.61 ± 0.04 & 88.40 ± 0.07 \\
    & Mixtral & 89.61 ± 0.08 & 88.38 ± 0.11 \\
    & Mixtral + TD & \textbf{89.65 ± 0.05} & \textbf{88.47 ± 0.05} \\
    \hline
    \multirow{6}{*}{NCN}
    & Gemini & 89.07 ± 0.25 & 87.46 ± 0.28 \\
    & Gemini + TD & \textbf{89.14 ± 0.27} & \textbf{87.75 ± 0.27} \\
    & LLaMA3 & 88.90 ± 0.24 & 87.33 ± 0.27 \\
    & LLaMA3 + TD & 89.08 ± 0.20 & 87.61 ± 0.19 \\
    & Mixtral & 88.99 ± 0.27 & 87.44 ± 0.25 \\
    & Mixtral + TD & 89.04 ± 0.24 & 87.51 ± 0.22 \\
    \hline
  \end{tabular}
  \caption{Link prediction performance using Time-Decayed (TD) embedding concatenation.}
  \label{table:td}
\end{table}

\section{Implications of Model Predictions: An Analytical Review by Domain Scientists}
Our domain scientists examined the connections commonly predicted correctly by the top three models. While many of the link predictions correspond to fundamental concept connections that have existed within the field of quantum information science for many years (e.g., “nonlinear oscillator” and “transmon,” “Hilbert space” and “quantum information,” etc.), some of the emerging connections within the test set seem timely and point towards recent trends and scientific breakthroughs within the field. Two examples relevant to quantum engineering that were observed within the data set are listed below.

The models accurately predicted a breakthrough in the coherent control of magnons \cite{Xu2023}. Methods for coupling classical magnons to photon cavities have been in development for the past decade \cite{Huebl2013, Tabuchi2015, Boventer2018, Boventer2020}, and the recent development of nonclassical coherent control of magnons built on this prior work. Therefore, it can be concluded that the models recognized the trend toward coherently controlled quantum magnonics for its prediction.

Likewise, the models recognized the importance of phonon engineering to the performance of superconducting qubits \cite{Kitzman2023}. This concept connection, which emerged naturally with the field of superconducting technologies, has been vitally important within the context of recent studies \cite{Wilen2021, Yelton2024} on gamma and muon ray impacts on superconducting quantum devices, wherein phonons serve as the mediating particle for qubit decoherence from such high-energy particles. Indeed, phonon engineering will likely prove to be an essential component of quantum engineering in the coming years.

\section{Related Work}
LLMs have shown impressive performance across numerous NLP tasks, particularly in node classification on graphs \citep{fatemi2024talk, chen2024labelfree}. However, they struggle to capture graph structural information \citep{wang2024can} and face scalability issues \citep{hu2020open} due to higher prediction costs compared to GNNs. Despite this, LLMs provide valuable semantic knowledge, particularly for node feature initialization, enhancing GNN performance in link prediction. In this study, we employed three advanced LLMs: Google's Gemini Pro \citep{team2023gemini}, a multimodal model for complex reasoning, Mixtral-8x7B \citep{jiang2024mixtral}, which supports long sequences and efficient inference, and Meta's Llama 3 \citep{metallama3}, known for its optimized architecture and versatility across tasks.

GNNs have become a powerful method for homogeneous link prediction. Architectures like GCN \citep{kipf2016semi}, GraphSAGE \citep{hamilton2017inductive}, and GAE \citep{kipf2016variational} encode node features and graph topology into low-dimensional embeddings for predicting link likelihood between nodes. Variational autoencoders (VAEs) \citep{ahn2021variational} further enhance representation learning by encoding data into a latent space for reconstruction. GNNs excel in capturing higher-order relationships and learning expressive node representations, outperforming traditional heuristic methods, especially in large, complex networks. Recent approaches like BUDDY \citep{chamberlain2023graph} and NCN \citep{wangneural} improve link prediction by leveraging pairwise information, including subgraphs and common neighbor data.

DeepWalk \citep{perozzi2014deepwalk}, LINE \citep{tang2015line}, and node2vec \citep{grover2016node2vec} are network embedding methods. DeepWalk applies Skip-Gram to node sequences generated by random walks, while LINE preserves local and global structures by optimizing first and second-order proximities. Node2vec introduces flexibility with biased random walks, interpolating between BFS and DFS using two parameters. These methods are essential for tasks such as link prediction, node classification, and recommendation systems.

\section{Conclusion and Future Work}

Our proposed approach offers a promising avenue for enhancing the performance of link prediction models, particularly in scenarios where initial node features are sparse or inadequate. This method not only enriches the feature set available for model training but also improves the model's ability to capture and represent complex patterns within the data. We applied this method to a quantum computing semantic network constructed from relevant scientific literature, and the models with node feature initialization by LLMs outperformed baseline node embedding methods across various link prediction models. Our approach is easily extendable to other graph datasets in different domains that lack adequate node features.

In this study, we focused exclusively on featurizing nodes within a graph, although edge features are also crucial for training models. Unlike node features, generating edge features via LLMs may not be practical due to the significantly higher number of edges compared to nodes. More effective edge feature generation methods by LLMs, such as clustering edges based on the characteristics of the involved nodes, will be explored in future research. Additionally, our experiments were limited to static graph settings. Dynamic GNNs and time-dependent graph methods could potentially improve prediction capabilities. Future work will aim to refine this approach further and explore its applicability in other graph-based learning tasks. 

\section*{Acknowledgments}

This material is based upon work conducted at Brookhaven National Laboratory (BNL) and supported by the Laboratory for Physical Sciences (LPS). BNL is operated and managed for the U.S. Department of Energy Office of Science by Brookhaven Science Associates under contract No. DE-SC0012704.

\bibliography{custom}

\begin{thebibliography}{53}
\providecommand{\natexlab}[1]{#1}

\bibitem[{Ahn and Kim(2021)}]{ahn2021variational}
Seong~Jin Ahn and MyoungHo Kim. 2021.
\newblock Variational graph normalized autoencoders.
\newblock In \emph{Proceedings of the 30th ACM international conference on information \& knowledge management}, pages 2827--2831.

\bibitem[{Arute et~al.(2019)Arute, Arya, Babbush, Bacon, Bardin, Barends, Biswas, Boixo, Brandao, Buell, Burkett, Chen, Chen, Chiaro, Collins, Courtney, Dunsworth, Farhi, Foxen, Fowler, Gidney, Giustina, Graff, Guerin, Habegger, Harrigan, Hartmann, Ho, Hoffmann, Huang, Humble, Isakov, Jeffrey, Jiang, Kafri, Kechedzhi, Kelly, Klimov, Knysh, Korotkov, Kostritsa, Landhuis, Lindmark, Lucero, Lyakh, Mandr{\`a}, McClean, McEwen, Megrant, Mi, Michielsen, Mohseni, Mutus, Naaman, Neeley, Neill, Niu, Ostby, Petukhov, Platt, Quintana, Rieffel, Roushan, Rubin, Sank, Satzinger, Smelyanskiy, Sung, Trevithick, Vainsencher, Villalonga, White, Yao, Yeh, Zalcman, Neven, and Martinis}]{Arute2019}
Frank Arute, Kunal Arya, Ryan Babbush, Dave Bacon, Joseph~C. Bardin, Rami Barends, Rupak Biswas, Sergio Boixo, Fernando G. S.~L. Brandao, David~A. Buell, Brian Burkett, Yu~Chen, Zijun Chen, Ben Chiaro, Roberto Collins, William Courtney, Andrew Dunsworth, Edward Farhi, Brooks Foxen, Austin Fowler, Craig Gidney, Marissa Giustina, Rob Graff, Keith Guerin, Steve Habegger, Matthew~P. Harrigan, Michael~J. Hartmann, Alan Ho, Markus Hoffmann, Trent Huang, Travis~S. Humble, Sergei~V. Isakov, Evan Jeffrey, Zhang Jiang, Dvir Kafri, Kostyantyn Kechedzhi, Julian Kelly, Paul~V. Klimov, Sergey Knysh, Alexander Korotkov, Fedor Kostritsa, David Landhuis, Mike Lindmark, Erik Lucero, Dmitry Lyakh, Salvatore Mandr{\`a}, Jarrod~R. McClean, Matthew McEwen, Anthony Megrant, Xiao Mi, Kristel Michielsen, Masoud Mohseni, Josh Mutus, Ofer Naaman, Matthew Neeley, Charles Neill, Murphy~Yuezhen Niu, Eric Ostby, Andre Petukhov, John~C. Platt, Chris Quintana, Eleanor~G. Rieffel, Pedram Roushan, Nicholas~C. Rubin, Daniel Sank, Kevin~J.
  Satzinger, Vadim Smelyanskiy, Kevin~J. Sung, Matthew~D. Trevithick, Amit Vainsencher, Benjamin Villalonga, Theodore White, Z.~Jamie Yao, Ping Yeh, Adam Zalcman, Hartmut Neven, and John~M. Martinis. 2019.
\newblock \href {https://doi.org/10.1038/s41586-019-1666-5} {Quantum supremacy using a programmable superconducting processor}.
\newblock \emph{Nature}, 574(7779):505--510.

\bibitem[{Berahmand et~al.(2021)Berahmand, Nasiri, Rostami, and Forouzandeh}]{berahmand2021modified}
Kamal Berahmand, Elahe Nasiri, Mehrdad Rostami, and Saman Forouzandeh. 2021.
\newblock A modified deepwalk method for link prediction in attributed social network.
\newblock \emph{Computing}, 103:2227--2249.

\bibitem[{Boventer et~al.(2020)Boventer, D\"orflinger, Wolz, Mac\^edo, Lebrun, Kl\"aui, and Weides}]{Boventer2020}
Isabella Boventer, Christine D\"orflinger, Tim Wolz, Rair Mac\^edo, Romain Lebrun, Mathias Kl\"aui, and Martin Weides. 2020.
\newblock \href {https://doi.org/10.1103/PhysRevResearch.2.013154} {Control of the coupling strength and linewidth of a cavity magnon-polariton}.
\newblock \emph{Phys. Rev. Res.}, 2:013154.

\bibitem[{Boventer et~al.(2018)Boventer, Pfirrmann, Krause, Sch\"on, Kl\"aui, and Weides}]{Boventer2018}
Isabella Boventer, Marco Pfirrmann, Julius Krause, Yannick Sch\"on, Mathias Kl\"aui, and Martin Weides. 2018.
\newblock \href {https://doi.org/10.1103/PhysRevB.97.184420} {Complex temperature dependence of coupling and dissipation of cavity magnon polaritons from millikelvin to room temperature}.
\newblock \emph{Phys. Rev. B}, 97:184420.

\bibitem[{Chamberlain et~al.(2023)Chamberlain, Shirobokov, Rossi, Frasca, Markovich, Hammerla, Bronstein, and Hansmire}]{chamberlain2023graph}
Benjamin~Paul Chamberlain, Sergey Shirobokov, Emanuele Rossi, Fabrizio Frasca, Thomas Markovich, Nils Hammerla, Michael~M Bronstein, and Max Hansmire. 2023.
\newblock Graph neural networks for link prediction with subgraph sketching.
\newblock \emph{ICLR}.

\bibitem[{Chen et~al.(2024)Chen, Mao, Wen, Han, Jin, Zhang, Liu, and Tang}]{chen2024labelfree}
Zhikai Chen, Haitao Mao, Hongzhi Wen, Haoyu Han, Wei Jin, Haiyang Zhang, Hui Liu, and Jiliang Tang. 2024.
\newblock \href {https://openreview.net/forum?id=hESD2NJFg8} {Label-free node classification on graphs with large language models ({LLM}s)}.
\newblock In \emph{The Twelfth International Conference on Learning Representations}.

\bibitem[{Cui et~al.(2022)Cui, Lu, Li, and Yang}]{cui2022positional}
Hejie Cui, Zijie Lu, Pan Li, and Carl Yang. 2022.
\newblock On positional and structural node features for graph neural networks on non-attributed graphs.
\newblock In \emph{Proceedings of the 31st ACM International Conference on Information \& Knowledge Management}, pages 3898--3902.

\bibitem[{Duong et~al.(2019)Duong, Hoang, Dang, Nguyen, and Aberer}]{duong2019node}
Chi~Thang Duong, Thanh~Dat Hoang, Ha~The~Hien Dang, Quoc Viet~Hung Nguyen, and Karl Aberer. 2019.
\newblock On node features for graph neural networks.
\newblock \emph{arXiv preprint arXiv:1911.08795}.

\bibitem[{Fatemi et~al.(2024)Fatemi, Halcrow, and Perozzi}]{fatemi2024talk}
Bahare Fatemi, Jonathan Halcrow, and Bryan Perozzi. 2024.
\newblock \href {https://openreview.net/forum?id=IuXR1CCrSi} {Talk like a graph: Encoding graphs for large language models}.
\newblock In \emph{The Twelfth International Conference on Learning Representations}.

\bibitem[{Gemini-Team et~al.(2023)Gemini-Team, Anil, Borgeaud, Wu, Alayrac, Yu, Soricut, Schalkwyk, Dai, Hauth et~al.}]{team2023gemini}
Gemini-Team, Rohan Anil, Sebastian Borgeaud, Yonghui Wu, Jean-Baptiste Alayrac, Jiahui Yu, Radu Soricut, Johan Schalkwyk, Andrew~M Dai, Anja Hauth, et~al. 2023.
\newblock Gemini: a family of highly capable multimodal models.
\newblock \emph{arXiv preprint arXiv:2312.11805}.

\bibitem[{Google-AI(2024)}]{google_text_embedding}
Google-AI. 2024.
\newblock Text embeddings api | generative ai on vertex ai.
\newblock \url{https://cloud.google.com/vertex-ai/generative-ai/docs/embeddings/get-text-embeddings}.

\bibitem[{Grover and Leskovec(2016)}]{grover2016node2vec}
Aditya Grover and Jure Leskovec. 2016.
\newblock node2vec: Scalable feature learning for networks.
\newblock In \emph{Proceedings of the 22nd ACM SIGKDD international conference on Knowledge discovery and data mining}, pages 855--864.

\bibitem[{Hamilton et~al.(2017)Hamilton, Ying, and Leskovec}]{hamilton2017inductive}
Will Hamilton, Zhitao Ying, and Jure Leskovec. 2017.
\newblock Inductive representation learning on large graphs.
\newblock \emph{Advances in neural information processing systems}, 30.

\bibitem[{Hu et~al.(2020)Hu, Fey, Zitnik, Dong, Ren, Liu, Catasta, and Leskovec}]{hu2020open}
Weihua Hu, Matthias Fey, Marinka Zitnik, Yuxiao Dong, Hongyu Ren, Bowen Liu, Michele Catasta, and Jure Leskovec. 2020.
\newblock Open graph benchmark: Datasets for machine learning on graphs.
\newblock \emph{Advances in neural information processing systems}, 33:22118--22133.

\bibitem[{Huang et~al.(2021)Huang, Lienhard, Calusine, Veps\"al\"ainen, Braum\"uller, Kim, Melville, Niedzielski, Yoder, Kannan, Orlando, Gustavsson, and Oliver}]{Huang2021}
Sihao Huang, Benjamin Lienhard, Greg Calusine, Antti Veps\"al\"ainen, Jochen Braum\"uller, David~K. Kim, Alexander~J. Melville, Bethany~M. Niedzielski, Jonilyn~L. Yoder, Bharath Kannan, Terry~P. Orlando, Simon Gustavsson, and William~D. Oliver. 2021.
\newblock \href {https://doi.org/10.1103/PRXQuantum.2.020306} {Microwave package design for superconducting quantum processors}.
\newblock \emph{PRX Quantum}, 2:020306.

\bibitem[{Huebl et~al.(2013)Huebl, Zollitsch, Lotze, Hocke, Greifenstein, Marx, Gross, and Goennenwein}]{Huebl2013}
Hans Huebl, Christoph~W. Zollitsch, Johannes Lotze, Fredrik Hocke, Moritz Greifenstein, Achim Marx, Rudolf Gross, and Sebastian T.~B. Goennenwein. 2013.
\newblock \href {https://doi.org/10.1103/PhysRevLett.111.127003} {High cooperativity in coupled microwave resonator ferrimagnetic insulator hybrids}.
\newblock \emph{Phys. Rev. Lett.}, 111:127003.

\bibitem[{Jiang et~al.(2024)Jiang, Sablayrolles, Roux, Mensch, Savary, Bamford, Chaplot, Casas, Hanna, Bressand et~al.}]{jiang2024mixtral}
Albert~Q Jiang, Alexandre Sablayrolles, Antoine Roux, Arthur Mensch, Blanche Savary, Chris Bamford, Devendra~Singh Chaplot, Diego de~las Casas, Emma~Bou Hanna, Florian Bressand, et~al. 2024.
\newblock Mixtral of experts.
\newblock \emph{arXiv preprint arXiv:2401.04088}.

\bibitem[{Jiang et~al.(2023)Jiang, Ren, and Lin}]{jiang2023llm}
Dongfu Jiang, Xiang Ren, and Bill~Yuchen Lin. 2023.
\newblock Llm-blender: Ensembling large language models with pairwise ranking and generative fusion.
\newblock In \emph{The 61st Annual Meeting Of The Association For Computational Linguistics}.

\bibitem[{Kamalloo et~al.(2023)Kamalloo, Dziri, Clarke, and Rafiei}]{kamalloo2023evaluating}
Ehsan Kamalloo, Nouha Dziri, Charles Clarke, and Davood Rafiei. 2023.
\newblock Evaluating open-domain question answering in the era of large language models.
\newblock In \emph{Proceedings of the 61st Annual Meeting of the Association for Computational Linguistics (Volume 1: Long Papers)}, pages 5591--5606.

\bibitem[{Kipf and Welling(2016{\natexlab{a}})}]{kipf2016semi}
Thomas~N Kipf and Max Welling. 2016{\natexlab{a}}.
\newblock Semi-supervised classification with graph convolutional networks.
\newblock \emph{arXiv preprint arXiv:1609.02907}.

\bibitem[{Kipf and Welling(2016{\natexlab{b}})}]{kipf2016variational}
Thomas~N Kipf and Max Welling. 2016{\natexlab{b}}.
\newblock Variational graph auto-encoders.
\newblock \emph{arXiv preprint arXiv:1611.07308}.

\bibitem[{Kitzman et~al.(2023)Kitzman, Lane, Undershute, Harrington, Beysengulov, Mikolas, Murch, and Pollanen}]{Kitzman2023}
J.~M. Kitzman, J.~R. Lane, C.~Undershute, P.~M. Harrington, N.~R. Beysengulov, C.~A. Mikolas, K.~W. Murch, and J.~Pollanen. 2023.
\newblock \href {https://doi.org/10.1038/s41467-023-39682-0} {Phononic bath engineering of a superconducting qubit}.
\newblock \emph{Nature Communications}, 14(1):3910.

\bibitem[{Kjaergaard et~al.(2020)Kjaergaard, Schwartz, Braumüller, Krantz, Wang, Gustavsson, and Oliver}]{Kjaergaard2020}
Morten Kjaergaard, Mollie~E. Schwartz, Jochen Braumüller, Philip Krantz, Joel I.-J. Wang, Simon Gustavsson, and William~D. Oliver. 2020.
\newblock \href {https://doi.org/10.1146/annurev-conmatphys-031119-050605} {Superconducting qubits: Current state of play}.
\newblock \emph{Annual Review of Condensed Matter Physics}, 11(Volume 11, 2020):369--395.

\bibitem[{Krantz et~al.(2019)Krantz, Kjaergaard, Yan, Orlando, Gustavsson, and Oliver}]{Krantz2019}
P.~Krantz, M.~Kjaergaard, F.~Yan, T.~P. Orlando, S.~Gustavsson, and W.~D. Oliver. 2019.
\newblock \href {https://doi.org/10.1063/1.5089550} {{A quantum engineer's guide to superconducting qubits}}.
\newblock \emph{Applied Physics Reviews}, 6(2):021318.

\bibitem[{Krenn and Zeilinger(2020)}]{krenn2020predicting}
Mario Krenn and Anton Zeilinger. 2020.
\newblock Predicting research trends with semantic and neural networks with an application in quantum physics.
\newblock \emph{Proceedings of the National Academy of Sciences}, 117(4):1910--1916.

\bibitem[{Li et~al.(2023)Li, Shomer, Mao, Zeng, Ma, Shah, Tang, and Yin}]{NEURIPS2023_0be50b45}
Juanhui Li, Harry Shomer, Haitao Mao, Shenglai Zeng, Yao Ma, Neil Shah, Jiliang Tang, and Dawei Yin. 2023.
\newblock \href {https://proceedings.neurips.cc/paper_files/paper/2023/file/0be50b4590f1c5fdf4c8feddd63c4f67-Paper-Datasets_and_Benchmarks.pdf} {Evaluating graph neural networks for link prediction: Current pitfalls and new benchmarking}.
\newblock In \emph{Advances in Neural Information Processing Systems}, volume~36, pages 3853--3866. Curran Associates, Inc.

\bibitem[{Martinis(2021)}]{Martinis2021}
John~M. Martinis. 2021.
\newblock \href {https://doi.org/10.1038/s41534-021-00431-0} {Saving superconducting quantum processors from decay and correlated errors generated by gamma and cosmic rays}.
\newblock \emph{npj Quantum Information}, 7(1):90.

\bibitem[{Meta-AI(2024)}]{metallama3}
Meta-AI. 2024.
\newblock Introducing meta llama 3: The most capable openly available llm to date.
\newblock \url{https://ai.meta.com/blog/llama3}.
\newblock Accessed: 2024-04-19.

\bibitem[{Montanaro(2016)}]{Montanaro2016}
Ashley Montanaro. 2016.
\newblock \href {https://doi.org/10.1038/npjqi.2015.23} {Quantum algorithms: an overview}.
\newblock \emph{npj Quantum Information}, 2(1):15023--15030.

\bibitem[{M{\"u}ller et~al.(2024)M{\"u}ller, Galkin, Morris, and Ramp{\'a}{\v{s}}ek}]{mullerattending}
Luis M{\"u}ller, Mikhail Galkin, Christopher Morris, and Ladislav Ramp{\'a}{\v{s}}ek. 2024.
\newblock Attending to graph transformers.
\newblock \emph{Transactions on Machine Learning Research}.

\bibitem[{Murray(2021)}]{Murray2021}
Conal~E. Murray. 2021.
\newblock \href {https://doi.org/10.1016/j.mser.2021.100646} {Material matters in superconducting qubits}.
\newblock \emph{Materials Science and Engineering: R: Reports}, 146:100646.

\bibitem[{Perozzi et~al.(2014)Perozzi, Al-Rfou, and Skiena}]{perozzi2014deepwalk}
Bryan Perozzi, Rami Al-Rfou, and Steven Skiena. 2014.
\newblock Deepwalk: Online learning of social representations.
\newblock In \emph{Proceedings of the 20th ACM SIGKDD international conference on Knowledge discovery and data mining}, pages 701--710.

\bibitem[{Place et~al.(2021)Place, Rodgers, Mundada, Smitham, Fitzpatrick, Leng, Premkumar, Bryon, Vrajitoarea, Sussman, Cheng, Madhavan, Babla, Le, Gang, J{\"a}ck, Gyenis, Yao, Cava, de~Leon, and Houck}]{Place2021}
Alexander P.~M. Place, Lila V.~H. Rodgers, Pranav Mundada, Basil~M. Smitham, Mattias Fitzpatrick, Zhaoqi Leng, Anjali Premkumar, Jacob Bryon, Andrei Vrajitoarea, Sara Sussman, Guangming Cheng, Trisha Madhavan, Harshvardhan~K. Babla, Xuan~Hoang Le, Youqi Gang, Berthold J{\"a}ck, Andr{\'a}s Gyenis, Nan Yao, Robert~J. Cava, Nathalie~P. de~Leon, and Andrew~A. Houck. 2021.
\newblock \href {https://doi.org/10.1038/s41467-021-22030-5} {New material platform for superconducting transmon qubits with coherence times exceeding 0.3 milliseconds}.
\newblock \emph{Nature Communications}, 12(1):1779.

\bibitem[{Rzhetsky et~al.(2015)Rzhetsky, Foster, Foster, and Evans}]{rzhetsky2015choosing}
Andrey Rzhetsky, Jacob~G Foster, Ian~T Foster, and James~A Evans. 2015.
\newblock Choosing experiments to accelerate collective discovery.
\newblock \emph{Proceedings of the National Academy of Sciences}, 112(47):14569--14574.

\bibitem[{Sedhain et~al.(2014)Sedhain, Sanner, Braziunas, Xie, and Christensen}]{sedhain2014social}
Suvash Sedhain, Scott Sanner, Darius Braziunas, Lexing Xie, and Jordan Christensen. 2014.
\newblock Social collaborative filtering for cold-start recommendations.
\newblock In \emph{Proceedings of the 8th ACM Conference on Recommender systems}, pages 345--348.

\bibitem[{Shor(1997)}]{Shor1997}
Peter~W. Shor. 1997.
\newblock \href {https://doi.org/10.1137/S0097539795293172} {Polynomial-time algorithms for prime factorization and discrete logarithms on a quantum computer}.
\newblock \emph{SIAM Journal on Computing}, 26(5):1484--1509.

\bibitem[{Tabuchi et~al.(2015)Tabuchi, Ishino, Noguchi, Ishikawa, Yamazaki, Usami, and Nakamura}]{Tabuchi2015}
Yutaka Tabuchi, Seiichiro Ishino, Atsushi Noguchi, Toyofumi Ishikawa, Rekishu Yamazaki, Koji Usami, and Yasunobu Nakamura. 2015.
\newblock \href {https://doi.org/10.1126/science.aaa3693} {Coherent coupling between a ferromagnetic magnon and a superconducting qubit}.
\newblock \emph{Science}, 349(6246):405--408.

\bibitem[{Tang et~al.(2015)Tang, Qu, Wang, Zhang, Yan, and Mei}]{tang2015line}
Jian Tang, Meng Qu, Mingzhe Wang, Ming Zhang, Jun Yan, and Qiaozhu Mei. 2015.
\newblock Line: Large-scale information network embedding.
\newblock In \emph{Proceedings of the 24th international conference on world wide web}, pages 1067--1077.

\bibitem[{Tang and Wang(2022)}]{tang2022cold}
Minghu Tang and Wenjun Wang. 2022.
\newblock Cold-start link prediction integrating community information via multi-nonnegative matrix factorization.
\newblock \emph{Chaos, Solitons \& Fractals}, 162:112421.

\bibitem[{Wang et~al.(2024{\natexlab{a}})Wang, Feng, He, Tan, Han, and Tsvetkov}]{wang2024can}
Heng Wang, Shangbin Feng, Tianxing He, Zhaoxuan Tan, Xiaochuang Han, and Yulia Tsvetkov. 2024{\natexlab{a}}.
\newblock Can language models solve graph problems in natural language?
\newblock \emph{Advances in Neural Information Processing Systems}, 36.

\bibitem[{Wang et~al.(2023)Wang, Yang, Huang, Yang, Majumder, and Wei}]{wang2023improving}
Liang Wang, Nan Yang, Xiaolong Huang, Linjun Yang, Rangan Majumder, and Furu Wei. 2023.
\newblock Improving text embeddings with large language models.
\newblock \emph{arXiv preprint arXiv:2401.00368}.

\bibitem[{Wang et~al.(2024{\natexlab{b}})Wang, Yang, and Zhang}]{wangneural}
Xiyuan Wang, Haotong Yang, and Muhan Zhang. 2024{\natexlab{b}}.
\newblock Neural common neighbor with completion for link prediction.
\newblock In \emph{The Twelfth International Conference on Learning Representations}.

\bibitem[{Wilen et~al.(2021)Wilen, Abdullah, Kurinsky, Stanford, Cardani, D'Imperio, Tomei, Faoro, Ioffe, Liu, Opremcak, Christensen, DuBois, and McDermott}]{Wilen2021}
C.~D. Wilen, S.~Abdullah, N.~A. Kurinsky, C.~Stanford, L.~Cardani, G.~D'Imperio, C.~Tomei, L.~Faoro, L.~B. Ioffe, C.~H. Liu, A.~Opremcak, B.~G. Christensen, J.~L. DuBois, and R.~McDermott. 2021.
\newblock \href {https://doi.org/10.1038/s41586-021-03557-5} {Correlated charge noise and relaxation errors in superconducting qubits}.
\newblock \emph{Nature}, 594(7863):369--373.

\bibitem[{Xu et~al.(2023)Xu, Gu, Li, Weng, Wang, Li, Wang, Zhu, and You}]{Xu2023}
Da~Xu, Xu-Ke Gu, He-Kang Li, Yuan-Chao Weng, Yi-Pu Wang, Jie Li, H.~Wang, Shi-Yao Zhu, and J.~Q. You. 2023.
\newblock \href {https://doi.org/10.1103/PhysRevLett.130.193603} {Quantum control of a single magnon in a macroscopic spin system}.
\newblock \emph{Phys. Rev. Lett.}, 130:193603.

\bibitem[{Yang et~al.(2015)Yang, Lichtenwalter, and Chawla}]{yang2015evaluating}
Yang Yang, Ryan~N Lichtenwalter, and Nitesh~V Chawla. 2015.
\newblock Evaluating link prediction methods.
\newblock \emph{Knowledge and Information Systems}, 45:751--782.

\bibitem[{Yelton et~al.(2024)Yelton, Larson, Iaia, Dodge, La~Magna, Baity, Pechenezhskiy, McDermott, Kurinsky, Catelani, and Plourde}]{Yelton2024}
E.~Yelton, C.~P. Larson, V.~Iaia, K.~Dodge, G.~La~Magna, P.~G. Baity, I.~V. Pechenezhskiy, R.~McDermott, N.~A. Kurinsky, G.~Catelani, and B.~L.~T. Plourde. 2024.
\newblock \href {https://doi.org/10.1103/PhysRevB.110.024519} {Modeling phonon-mediated quasiparticle poisoning in superconducting qubit arrays}.
\newblock \emph{Phys. Rev. B}, 110:024519.

\bibitem[{Zanardini and Serrano(2024)}]{zanardini2024introducing}
Damiano Zanardini and Emilio Serrano. 2024.
\newblock Introducing new node prediction in graph mining: Predicting all links from isolated nodes with graph neural networks.
\newblock \emph{CoRR}.

\bibitem[{Zhang and Chen(2018)}]{zhang2018link}
Muhan Zhang and Yixin Chen. 2018.
\newblock Link prediction based on graph neural networks.
\newblock \emph{Advances in neural information processing systems}, 31.

\bibitem[{Zhang and Wang(2015)}]{zhang2015collective}
Wei Zhang and Jianyong Wang. 2015.
\newblock A collective bayesian poisson factorization model for cold-start local event recommendation.
\newblock In \emph{Proceedings of the 21th ACM SIGKDD international conference on knowledge discovery and data mining}, pages 1455--1464.

\bibitem[{Zhao et~al.(2017)Zhao, Du, and Buntine}]{zhao2017leveraging}
He~Zhao, Lan Du, and Wray Buntine. 2017.
\newblock Leveraging node attributes for incomplete relational data.
\newblock In \emph{International conference on machine learning}, pages 4072--4081. PMLR.

\bibitem[{Zhu et~al.(2023)Zhu, Yuan, Wang, Liu, Liu, Deng, Chen, Dou, and Wen}]{LLM4IRSurvey}
Yutao Zhu, Huaying Yuan, Shuting Wang, Jiongnan Liu, Wenhan Liu, Chenlong Deng, Haonan Chen, Zhicheng Dou, and Ji-Rong Wen. 2023.
\newblock \href {https://arxiv.org/abs/2308.07107} {Large language models for information retrieval: A survey}.
\newblock \emph{CoRR}, abs/2308.07107.

\bibitem[{Zhu et~al.(2021)Zhu, Zhang, Xhonneux, and Tang}]{zhu2021neural}
Zhaocheng Zhu, Zuobai Zhang, Louis-Pascal Xhonneux, and Jian Tang. 2021.
\newblock Neural bellman-ford networks: A general graph neural network framework for link prediction.
\newblock \emph{Advances in Neural Information Processing Systems}, 34:29476--29490.

\end{thebibliography}

\appendix

\section{Quantum Computing related Papers in arXiv}
\label{sec:appendix-pub-stat}

\begin{figure}
\centering
\begin{tikzpicture}
\begin{axis}[
    x tick label style={rotate=45, anchor=east, font=\small, /pgf/number format/set thousands separator={}}, 
    y tick label style={font=\small}, 
    xmin=2007, xmax=2024,
    ymin=0, ymax=16000,
    xtick=data,
    nodes near coords, 
    nodes near coords align={vertical}, 
    every node near coord/.append style={font=\tiny, below}, 
    ytick={0,2000,4000,6000,8000,10000,12000,14000,16000},
    ymajorgrids=true,
    grid=major, 
    grid style={dashed,gray!30}, 
]
 
\addplot[
    blue,
    mark=*
]
coordinates {
    (2024,9922)
    (2023,13529)
    (2022,10420)
    (2021,9524)
    (2020,8308)
    (2019,7143)
    (2018,6668)
    (2017,6619)
    (2016,7308)
    (2015,13156)
    (2014,4803)
    (2013,5043)
    (2012,3595)
    (2011,3422)
    (2010,2744)
    (2009,12632)
    (2008,2595)
    (2007,8691)
};

\end{axis}
\end{tikzpicture}
\caption{The number of quantum computing related papers in arXiv from 2007 to 2024 (as of June 15, 2024)}
\label{fig:paper-stat}
\end{figure}
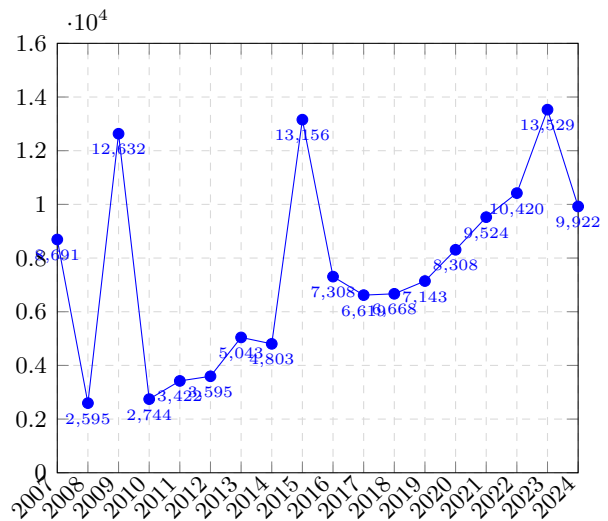

Figure~\ref{fig:paper-stat} illustrates the number of quantum computing papers published on arXiv over time.

\section{Evaluation on Isolated Nodes}
\label{sec:appendix-eval-iso}
Table~\ref{table:eval-iso} presents the model evaluation on the 1,368 connections to the isolated nodes.

\begin{table*}
  \centering
  \renewcommand{\arraystretch}{1.3} 
  \small
  \begin{tabular}{l|cccccc}
    \hline
    \multirow{2}{*}{\textbf{Node Embedding}} & \multicolumn{2}{c}{\textbf{MLP}} & \multicolumn{2}{c}{\textbf{GCN}} & \multicolumn{2}{c}{\textbf{GraphSAGE}} \\
    & AUROC & AP & AUROC & AP & AUROC & AP \\
    \hline
    DeepWalk & 71.57 ± 1.33 & 49.80 ± 1.96 & 49.18 ± 2.16 & 19.95 ± 1.60 & 53.26 ± 1.84 & 22.18 ± 1.85 \\
    LINE & 69.53 ± 5.64 & 50.06 ± 6.79 & \textbf{87.01 ± 0.10} & \textbf{68.21 ± 0.72} & 65.91 ± 11.62 & 39.25 ± 15.08 \\
    node2vec & 53.28 ± 3.40 & 23.40 ± 4.72 & 51.80 ± 0.71 & 21.34 ± 0.91 & 49.67 ± 0.22 & 19.31 ± 0.01 \\
    \hline
    Gemini-1.0-pro & 81.37 ± 1.16 & 62.91 ± 1.10 & 58.40 ± 9.04 & 28.42 ± 8.27 & 70.70 ± 4.63 & 40.85 ± 7.63 \\
    LLaMA3 (70B) & 84.79 ± 1.09 & 65.09 ± 2.78 & 56.06 ± 12.38 & 27.67 ± 10.48 & 72.54 ± 3.49 & 42.58 ± 4.42 \\
    Mixtral-8x7B (46B) & \textbf{85.55 ± 1.43} & \textbf{69.02 ± 1.91} & 51.28 ± 13.48 & 23.88 ± 10.68 & \textbf{75.92 ± 4.01} & \textbf{45.25 ± 5.38} \\
    \hline
  \end{tabular}
  \\[0.3cm]
  \begin{tabular}{l|cccccc}
    \hline
    \multirow{2}{*}{\textbf{Node Embedding}} & \multicolumn{2}{c}{\textbf{GAE}} & \multicolumn{2}{c}{\textbf{NCN}} & \multicolumn{2}{c}{\textbf{BUDDY}} \\
    & AUROC & AP & AUROC & AP & AUROC & AP \\
    \hline
    DeepWalk & 68.82 ± 3.50 & 42.41 ± 4.90 & 29.81 ± 18.98 & 17.45 ± 15.45 & 76.56 ± 3.50 & 48.86 ± 4.84 \\
    LINE & 86.95 ± 0.09 & 68.48 ± 0.74 & 60.85 ± 1.60 & 38.64 ± 1.04 & 80.09 ± 3.44 & 52.26 ± 3.01 \\
    node2vec & 36.34 ± 8.40 & 15.16 ± 2.82 & 42.40 ± 11.97 & 17.23 ± 5.38 & 79.59 ± 1.16 & 52.99 ± 3.47 \\
    \hline
    Gemini-1.0-pro & \textbf{89.23 ± 0.24} & \textbf{70.99 ± 0.72} & \textbf{69.96 ± 13.01} & \textbf{39.05 ± 15.39} & 85.17 ± 2.40 & 58.19 ± 2.80 \\
    LLaMA3 (70B) & 88.95 ± 0.37 & 70.01 ± 1.22 & 69.23 ± 16.72 & 40.42 ± 17.79 & \textbf{85.68 ± 2.26} & \textbf{57.57 ± 2.44} \\
    Mixtral-8x7B (46B) & 88.67 ± 0.38 & 70.64 ± 0.81 & 67.11 ± 15.40 & 38.10 ± 17.71 & 83.68 ± 4.50 & 55.82 ± 5.11 \\
    \hline
  \end{tabular}
  \caption{Comparison of LLM-generated node embeddings with other node embeddings in link prediction methods on 1,382 edges to 30 isolated nodes in the quantum computing concept graph.}
  \label{table:eval-iso}
\end{table*}

\section{Time Decay Embedding Generation}
\label{sec:appendix-td}

Time-decayed information is essential for analyzing time series data and predicting future trends due to several reasons: First, it ensures the relevance of recent data, which is often more indicative of future behavior than older data. This is crucial in scenarios like stock market analysis where recent trends are more predictive. Second, it allows models to adapt quickly to changes by emphasizing recent data, which reflects current underlying processes more accurately. Third, time decay reduces noise by minimizing the influence of older, potentially irrelevant data, thus focusing on meaningful patterns. Fourth, it enhances computational efficiency by potentially discarding less important older data. Finally, time-decayed information aids in anomaly detection by highlighting recent unusual behaviors. In summary, time-decayed information enables models to focus on the most pertinent data, adapt to changes, reduce noise, improve efficiency, and identify anomalies, thus providing a robust tool for time series analysis and trend prediction.

We utilized time decay information of pairs of concepts over time to represent node embeddings. To this end, we created co-occurrence matrices by each year. We then converted the co-occurrence matrices to PPMI (Positive Point-wise Mutual Information) matrices that are useful in word-word co-occurrence matrices as it addresses issues of normalization, sparsity, and noise reduction, thereby enhancing the quality and utility of semantic representations derived from such matrices. We adopted an exponential time decay function that assigns decreasing weights or importance to past events or observations based on their age or distance from the present. In the context of time series data or decay processes, an exponential decay function is commonly expressed as: $N(t) = N_0 e^{-\lambda t}$ where: $t$ is the time elapsed since the event or observation. $\lambda$ is a decay constant that determines how quickly the weight decreases over time. After applying the time decay function, the matrices were aggregated and then the dimension of the aggregated matrix was reduced to the same embedding size to the LLM feature embedding by the SVD (Singular Value Decomposition). This time decayed embeddings were concatenated with the LLM feature embeddings. Algorithm~\ref{alg:td} illustrates the node embedding generation procedure.

\begin{algorithm}
\caption{Time-Decayed Node Embedding Generation Procedure}
\begin{algorithmic}[1]
\REQUIRE Co-occurrence matrices for each year
\ENSURE Node embeddings
\STATE Convert co-occurrence matrices to PPMI matrices
\FOR{each year}
\STATE Apply exponential time decay function: $N(t) = N_0 e^{-\lambda t}$
\ENDFOR
\STATE Aggregate the matrices
\STATE Reduce the dimension of the aggregated matrix to the same embedding size as the LLM feature embedding using SVD
\STATE Concatenate the time decayed embeddings with the LLM feature embeddings
\end{algorithmic}
\label{alg:td}
\end{algorithm}

\end{document}